# Detection, segmentation and recognition of Face and its features using neural network.

Smriti Tikoo[1], Nitin Malik[2]
Research Scholar, Department of EECE, The Northcap University, Gurgaon, India.
Associate Professor, Department of EECE, The Northcap University, Gurgaon, India
Email: smrititikoo@gmail.com, nitinmalik@ncuindia.edu

**ABSTRACT:** Face detection and recognition has been prevalent with research scholars and diverse approaches have been incorporated till date to serve purpose. The rampant advent of biometric analysis systems, which may be full body scanners, or iris detection and recognition systems and the finger print recognition systems, and surveillance systems deployed for safety and security purposes have contributed to inclination towards same. Advances has been made with frontal view, lateral view of the face or using facial expressions such as anger, happiness and gloominess, still images and video image to be used for detection and recognition. This led to newer methods for face detection and recognition to be introduced in achieving accurate results and economically feasible and extremely secure. Techniques such as Principal Component analysis (PCA), Independent component analysis (ICA), Linear Discriminant Analysis (LDA), have been the predominant ones to be used. But with improvements needed in the previous approaches Neural Networks based recognition was like boon to the industry. It not only enhanced the recognition but also the efficiency of the process. Choosing Backpropagation as the learning method was clearly out of its efficiency to recognize non linear faces with an acceptance ratio of more than 90% and execution time of only few seconds.

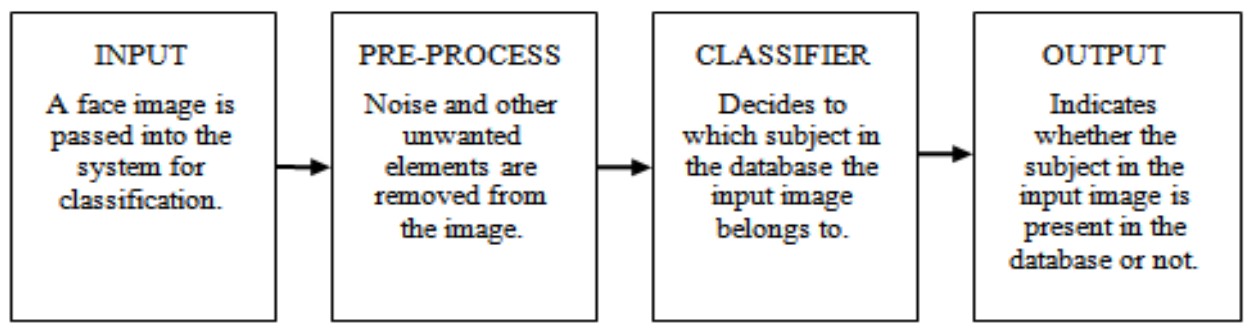

Figure 1: Face detection process.

## INTRODUCTION

Automatic recognition dates back to the years of 1960's when pioneers such as Woody Bledsoe, Helen Chan Wolf, and Charles Bisson introduced their works to the world. In 1964 & 1965, Bledsoe, along with Helen Chan and Charles Bisson had worked on the computer to recognize human faces. But it didn't allow much recognition to his work due to not much support from agencies. After him the work was carried forward by Peter Hart at the Stanford Research Institute. The results given by computer were terrific when feeded with a gallery of images for recognition. In around 1997, another approach by Christoph von der Malsburg and students of lot of reputed universities got together to design a system which was even better than the previous one. The system designed was robust and could identify people from photographs which didn't offer a clear view of the facial images .After that technologies like Polar and FaceIt were introduced which were doing the work in lesser time and gave accurate results by offering features like a 3D view of the picture and comparing it to a database from worldwide. Till the year 2006 a lot of new algorithm replaced the previous ones and were proving to be better in terms of offering various new techniques to get better results and their performance was commendable. Automatic Recognition has seen a lot of improvement since 1993, the error rate has dropped down & enhancement methods to improve the resolution have hit the market. Ever since then biometric authentication is all over the place, these days ranging from home security systems, surveillance cameras, to fingerprint and iris authentication systems, user verification and etc. These systems have contributed in catching terrorists and other type of intruders via cyber or in person. Recognition of faces has been under a research since 1960's but it's just over a decade that acceptable results have been obtained. Approaches deployed do not count the intricate details which differ from person to person in context of recognition results. The continuous changes in expression of face depending upon the situation being experienced by person disclose important information regarding the structural characteristics and the texture, thereby helping in recognition process. Of the many methods adopted towards recognition of face they can be broadly classified as a) holistic approach which relies of taking the whole face as an input in pixels and b) feature extraction which concentrates on capturing the features in a face but certain methods are deployed for removing the redundancy information and dimension problems involved. Three basic approaches in this direction are namely PCA, LDA, ICA others which were added to the list include the neural network, fuzzy and skin and texture based approaches also including the ASM model. The problem is approached first by detecting the face and its features and then moving on to recognition. A lot approaches are available for detection of faces like Viola Jones algorithm, LBP, Adaboost algorithm, neural network based detection or usage of Snow classifier method .

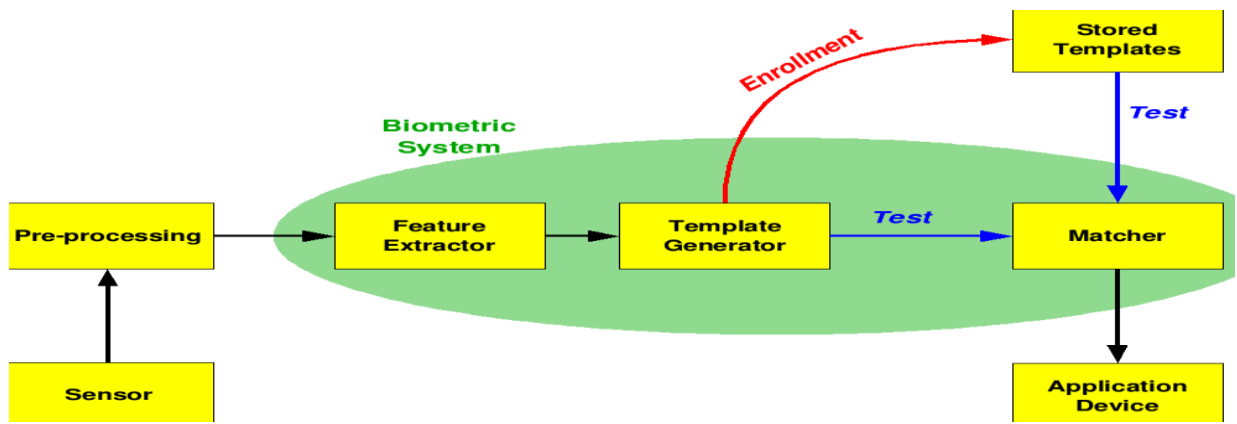

Figure 2: A generalized face recognition procedure

**PROBLEM DESCRIPTION**

The onset of biometric authentication dates back to 1960's but has seen a boost in the recent years due to constant threat to security from intruders via cyber or via crevices of the inhabited areas of country. It has found usage in security systems installed in homes, retina scans, hand scans, facial scans for verifications at airports or organizations or malls which are frequently visited areas by people of different races and culture. These systems tend to offer a sense of reliability, accountability and security from the threat at doorstep. It manages to safe guard people's interests and keeps the threats at bay as much as possible. This has brought the human computer interaction to a higher level, where computer are trying to imitate the functions performed by brain with the help of sensory organs. These tasks are performed with ease by our brains due to networking of neuron structure built in our brain in way a mechanism evolved to cope up with responding appropriately via motor and coordination skills. Such systems are mastered using various available approaches which make use of image processing and open source computer vision in such a way that have boosted the business for detection and recognition systems. In this paper we propose a similar approach to detect and recognize a facial image and its features using a BPNN with help of Matlab. In the Matlab we have worked under the neural network, using its tools to train and process the image for obtaining the performance and regression plots.

**BACKPROPAGATION:**

Also called as "backward propagation of errors", is a popular method of training MLP artificial neural networks in conjunction with an optimization method. Gradient of loss function is calculated with respect to the associated weights. Optimization method is fed with the gradient which alters the weights used to reduce the loss function. This algorithm works on the supervised learning rule i.e. requires a known, desired output for each input value in order to calculate the loss function gradient.It makes use of sigmoid function as its activation usually. The algorithm can be split in two phases:

Phase 1: **Propagation**:
**Forward propagation**: Input is fed through the network to generate propagation's output activations.
**Backward propagation**: A feedback network is formed by feeding the output as input in order to generate a difference between actual and the target outputs.

Phase 2: **Weight update**:
Gradient of weight is a product of difference of outputs and input activation.
Subtract a ratio (percentage) of the gradient from the weight.
Ratio or the learning rate affects the speed and the quality of learning. Greater the ratio the neurons are trained quickly but accuracy is assured with a lower ratio. If weight updation increases in positive side the error shall increase. Repeat the steps to attain satisfactory results.

**NEURAL NETWORK**:

Defines a family of algorithms via a multiple layers of interconnected inspired from the neuron structure in the brain. Here each circular node represents an artificial neuron and an arrow represents the connection from the output of one neuron to the input of another. These networks are designed for approximating functions for huge amount of inputs generally unknown. The nodes in neural network exchange data between one another. The connections between the nodes have numeric's weights associated which are altered to attain an output matching the target output in ace of supervised learning or making neural nets adaptive to inputs and capable of learning.

On the basis of the following criteria a network can be classified as neural: if it has set of adaptive weights and is capable of approximating nonlinear functions of their inputs. The adaptive weights can be thought of as connection strengths between neurons which are activated during training and predictions
.
**FACE DETECTION ALGORITHM**:

Proposed by Paul Viola and Michael Jones, it proves to be a one of its kind by providing competitive results for real time situations .It not only aces the process of detecting frontal faces but can also be modeled for detecting variety of other object classes. It's robust nature and its adaptability for practical situation makes its popular among its users. It follows a four stage procedure for the entire face detection which is: first is Haar feature selection, then comes the integral image creation followed by adaboost training and cascading of amplifiers. Haar Feature selection matches the commonalities found in human faces. The integral image calculates the rectangular features in fixed time which benefits it over other sophisticated features. Integral image at (x, y) coordinates gives the pixel sum of the coordinates above and on to the left of the (x,y).Ada boost training algorithm boost the perfor-mance of cascading amplifiers by training them appropriately so as to form a strong classifier. The strong classifiers are arranged in a cascade in order of complexity, where each successive classifier is trained only on those selected samples which pass through the preceding classifiers. If at any stage in the cascade a classifier rejects the sub-window under inspection, no further processing is performed and continue on searching the next sub-window. The cascade therefore has the form of a degenerate tree. In the case of faces, the first classifier in the cascade – called the attentional operator – uses only two features to achieve a false negative rate of approximately 0% and a false positive rate of 40%.The effect of this single classifier is to reduce by roughly half the number of times the entire cascade is evaluated. In cascading, each stage consists of a strong classifier. So all the features are grouped into several stages where each stage has certain number of features. The job of each stage is to determine whether a given sub-window is definitely not a face or may be a face. A given sub-window is immediately discarded as not a face if it fails in any of the stages

**VISION CASCADE OBJECT DETECTOR**:

Detect objects using the Viola-Jones algorithm The cascade object detector uses the Viola-Jones algorithm to detect people's faces, noses, eyes, mouth, or upper body. One can also use the Training Image Labeler to train a custom classifier to use with this System object.

**COMPUTER VISION TOOLS**: provides algorithms, functions, and apps for designing and simulating computer vision and video processing systems. You can perform feature detection, extraction, and matching; object detection and tracking; motion estimation; and video processing. For 3-D computer vision, the system toolbox supports camera calibration, stereo vision, 3-D reconstruction, and 3-D point cloud processing. With machine learning based frameworks, you can train object detection, object recognition, and image retrieval systems. Algorithms are available as MATLAB functions, System objects, and Simulink blocks. For rapid prototyping and embedded system design, the system toolbox supports fixed-point arithmetic and C-code generation.

**PROPOSED METHODOLOGY:**

1. Detection of face and its major features.
2. Capturing Images to be recognized in rgb, gray and binary format.
3. Compute histogram for the same.
4. Carry out the segmentation process of the images in rgb format.
5. Compute the histogram for them.
6. Compute histogram data into neural network fitting tool
7. Obtain the performance and regression plots for the same

**RESULTS AND DISCUSSION**:

After detecting faces and its features we move on the task of recognition using the Neural Network training for recognition to be completed. Here I am using different faces to show the procedure of face recognition one can continue with the same. So first we read the image convert to grayscale or binary as we will be making use of their equivalent histogram for feeding the data in the neural network. After opening the nft tool we choose the data to be trained for recognition procedure. That data is usually imported from an excel book. After importing the data neural network tool starts the training, It carries out training in 3 states that is training validation and testing. Where in the training stage the network is adjusted as per the error generated during it. During the validation network generalization is monitored and is used to halt the training .In testing the final solution is tested to confirm the predictive power of the network. Usually a lot of training sessions are conducted to enhance the performance of network and lower the mean squared error value which is the average squared difference between output and target. We have used feed forward networks under the supervised learning architecture of neural network tool box to compute our data, in which one way connections operate and no feedback mechanism is included. In testing the final solution was obtained by carrying out the training session for 5-10 times in order to achieve better outcomes .Mean squared error was the parameter chosen to assess the network's execution so far. Lower values for mean squared were obtained for the same. Thus with results one can conclude to have recognized the given image. The accuracy of this network is 85% approx.

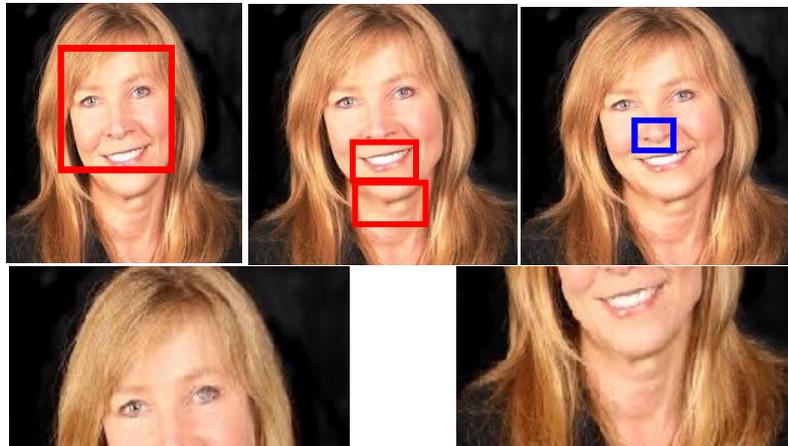

Figure 1: Detection of face and its features followed by the segmentation process.

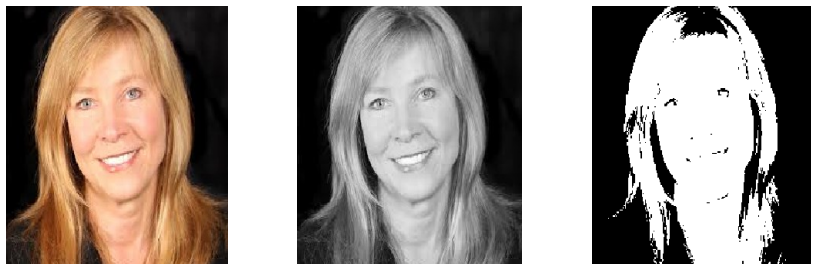

Figure2:The three format of sample image rgb, gray and binary

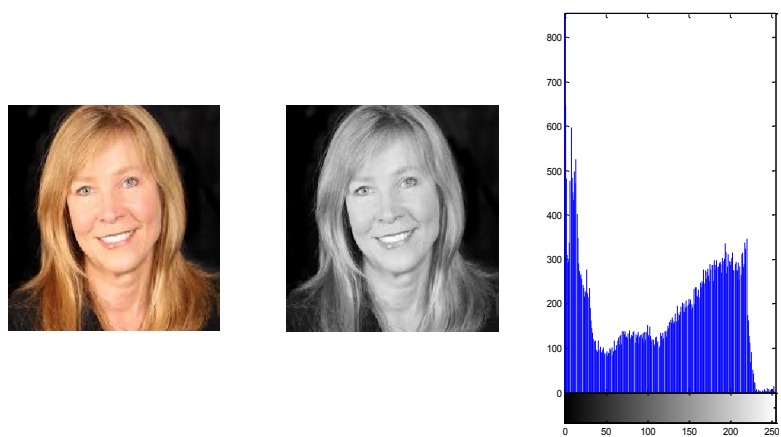

Figure3: Computation of histogram equivalent for the gray scale of the sample image.

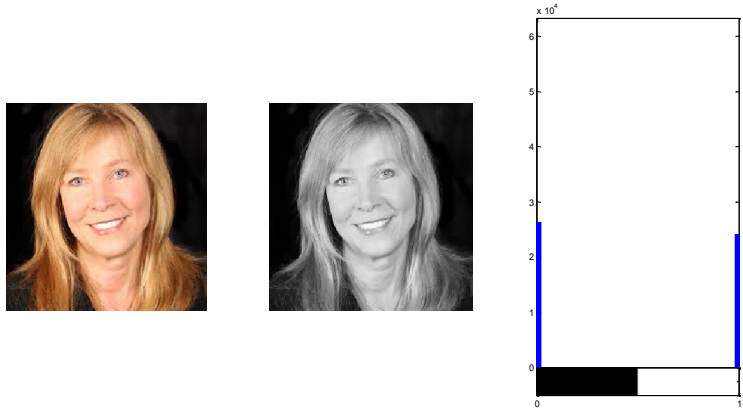

Figure4:Computation of binary histogram equivalent

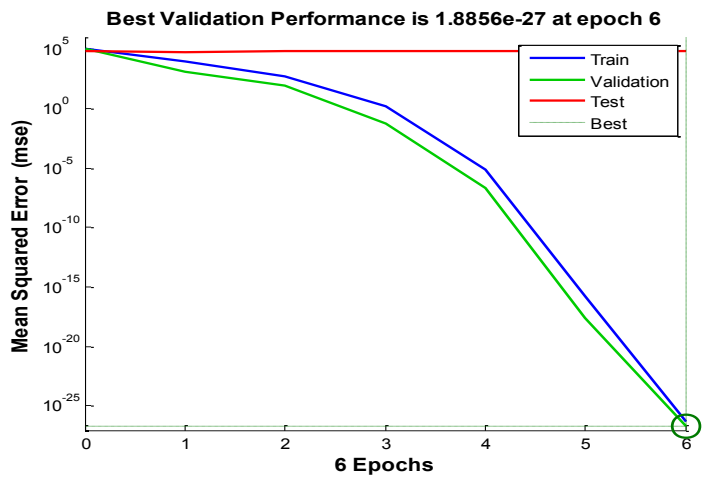

Figure5: Performance plot for grayscale image

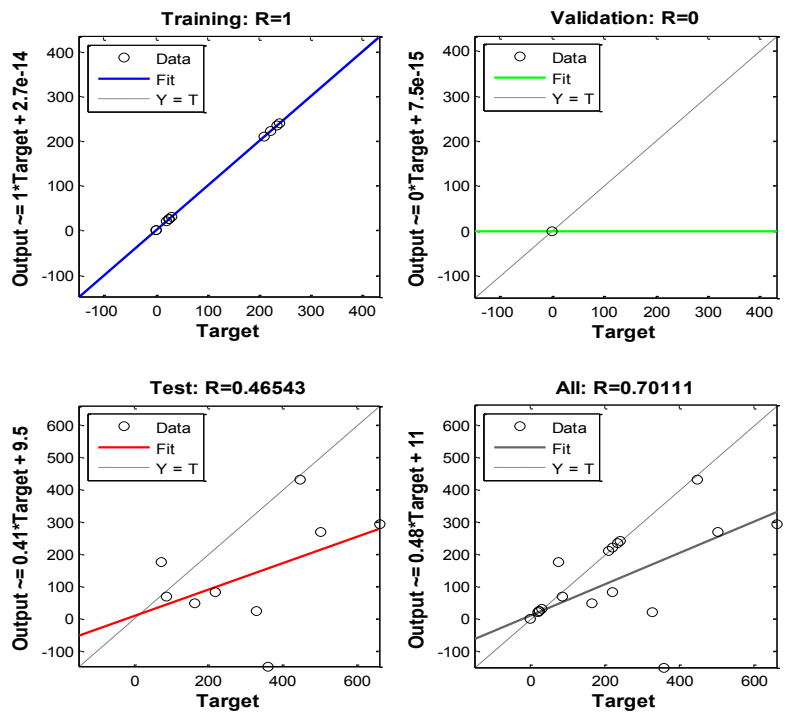

Figure 6:Regression plot for grayscale image

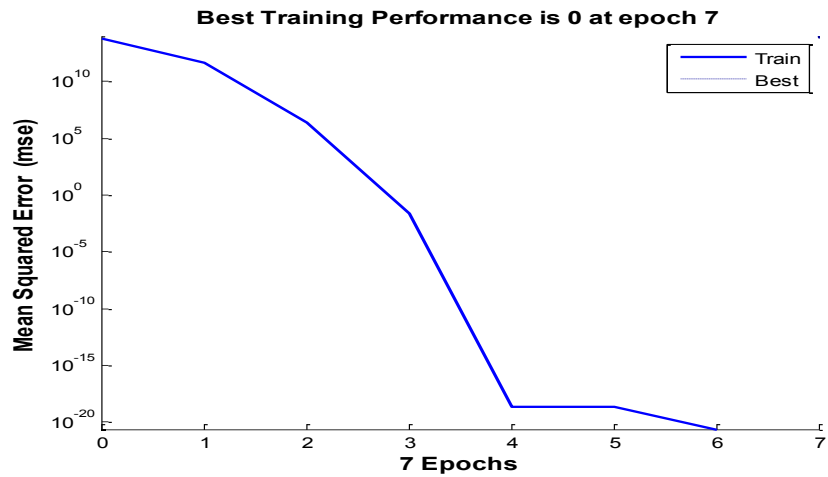

Figure 7:Performance plot for binary image

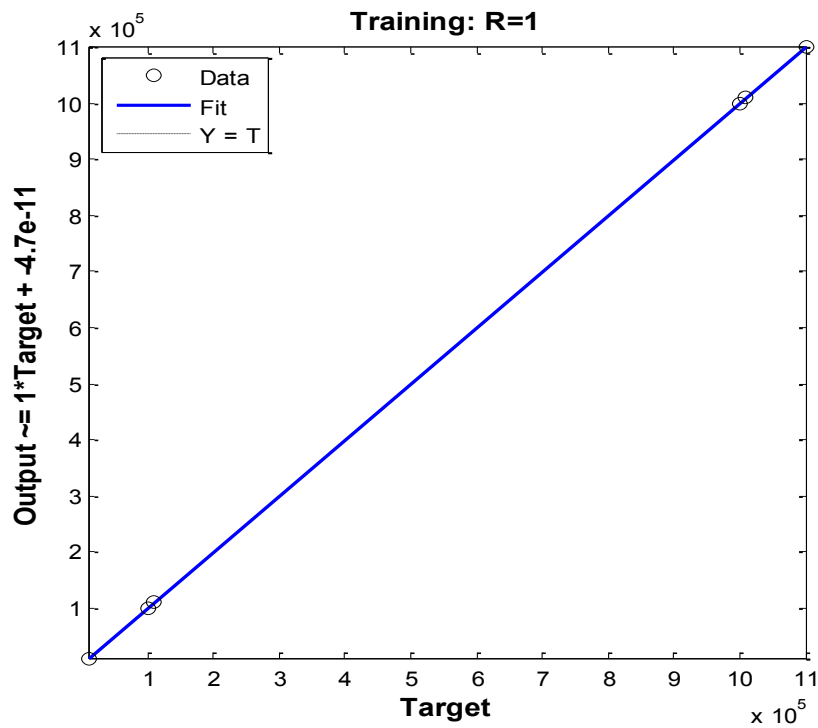

Figure 8: Regression plot for binary image

**CONCLUSION:**

In this work it has been shown that if a facial image of a person is given then the network can able to recognize the face of the person. The whole work is completed through the following steps: Facial image of a person has been collected by taking three different samples of the person for the experiment. Each image is divided into two equal parts to show the process of segmentation. Feed Forward Back Propagation neural network have been used to train, test and validate the net-work for each part of the image using MATLAB. The histogram equivalent of binary and gray images have been computed as the data to check for the recognition process using performance plot, regression plot as means of parameters to check the performance. Training of each sample is performed up to 3-4 times to minimize the error and changing the number of neurons in the hidden layer accordingly to obtain a better result. Facial image without dividing into parts have also been applied in the network.

**REFRENCES**

[1] H.A. Rowley, S. Baluja, and T. Kanade, "Neural network - based face detection," IEEE Trans. Pattern Analysis and Machine Intelligence, vol. 20, no. 1, pp. 23–38, Jan. 1998.

[2] H.A.Rowley, S. Baluja, and T. Kanade, "Rotation invariant neural net-workbased face detection," Proc. IEEE Int'l Conf. Computer Vision and Pattern Recognition, pp. 38–44, 1998.

[3] K.K Sung and T. Poggio, "Example-based learning for view-based human face detection," IEEE Trans. Pattern Analysis and Machine Intelli-gence, vol. 20, no. 1, pp. 39–51, Jan. 1998.

[4] H. Schneiderman and T. Kanade, "A statistical method for 3D object detection applied to faces and cars," Proc. IEEE Int'l Conf. Computer Vision and Pattern Recognition, pp. 746–751, June 2000.

[5] K.C. Yow and R. Cipolla, "Feature-based human face detection," Image and Vision Computing, vol. 25, no. 9, pp. 713–735, Sept. 1997.

[6] D. Maio and D. Maltoni, "Real-time face location on gray-scale static images," Pattern Recognition, vol. 33, no. 9, pp. 1525–1539, Sept. 2000.

[7] M.S. Lew and N. Huijsmans, "Information theory and face detection," Proc.IEEE Int'l Conf. Pattern Recognition, pp. 601- 605, Aug. 1996.

[8] S.C. Dass and A.K. Jain, "Markov face models," Proc. IEEE Int'l Conf.Computer Vision, pp. 680–687, July 2001.

[9] A.J. Colmenarez and T.S. Huang, "Face detection with information based maximum discrimina-tion," Proc. IEEE Int'l Conf. Computer Vision and Pattern Recognition, pp. 782–787, June 1997.

[10] D. DeCarlo and D. Metaxas, "Optical flow constraints on deformable models with applications to face tracking," International Journal Computer Vision, vol. 38, no. 2, pp. 99–127, July 2000.

[11] V. Bakic and G. Stockman, "Menu selection by facial aspect," Proc. Vision Interface, Canada, pp. 203–209, May 1999.

[12] A. Colmenarez, B. Frey, and T. Huang, "Detection and tracking of faces and facial features," Proc. IEEE Int'l Conf. Image Processing, pp. 657–661, Oct. 1999.

[13] R. F´eraud, O.J. Bernier, J.-E. Viallet, and M. Collobert, "A fast and accurate face detection based on neural network," IEEE Trans. Pattern Analysis and Machine Intelligence, vol. 23.

[14] W.Zhao, R.Chellappa, P.J..Phillips and A. Rosennfeld, "Face Reconi-tion: A literature Survey". ACM Comput.Surv., 35(4): 399-458, 2003.

[15] M.A.Turk and A.P. Pentaland, "Face Recognition Using Eigenfaces", IEEE conf. on Computer Vision and Pattern Recognition, pp. 586-591, 1991.

[16] Ling-Zhi Liao, Si-Wei Luo, and Mei Tian ""Whitenedfaces" Recogni-tion With PCA and ICA" IEEE Signal Processing Letters, Vol. 14, No. 12, pp1008-1011, Dec. 2007.

[17] G. Jarillo, W.Pedrycz , M. Reformat "Aggregation of classifiers based on image transformations in biometric face recognition" Machine Vision and Applications (2008) Vol . 19,pp. 125-140, Springer-Verlag 2007.

[18]  Tej Pal Singh, "Face Recognition by using Feed Forward Back Propagation Neural Network", International Journal of Innovative Research in Technology & Science, vol.1, no.1.

[19]  N.Revathy, T.Guhan, "Face recognition system using back propagation artificial neural networks", International Journal of Advanced Engineering Technology, vol.3, no. 1, 2012.

[20] Kwan-Ho Lin, Kin-Man Lam, and Wan-Chi Siu. "A New Approach using ModiGied Hausdorff Distances with EigenFace for Human Face Recognition" IEEE Seventh international Conference on Control, Automation, Robotics and Vision , Singapore, 2-5 Dec 2002, pp 980-984.

[21]  Zdravko Liposcak,  Sven Loncaric, "Face Recognition From Profiles Using Morphological Operations", IEEE Conference on Recognition, Analysis, and Tracking of Faces and Gestures in Real-Time Systems, 1999.

[22]  Simone Ceolin , William A.P Smith, Edwin Hancock, "Facial Shape Spaces from Surface Normals and Geodesic Distance", Digital Image Computing Techniques and Applications, 9$^{th}$ Biennial Conference of Australian Pattern Recognition Society IEEE 3-5 Dec.,2007 ,Glenelg, pp-416-423.

[23]  Ki-Chung Chung , Seok Cheol Kee ,and Sang Ryong Kim, "Face Recognition using Principal Component Analysis of Gabor Filter Responses" ,Recognition, Analysis, and Tracking of Faces and Gestures in Real-Time Systems, 1999,Proceedings. International Workshop IEEE  26-27 September,1999, Corfu,pp-53-57.

[24]  Ms. Varsha Gupta, Mr. Dipesh Sharma, "A Study of Various Face Detection Methods", International Journal of Advanced Research in Computer and Communication Engineering), vol.3, no. 5, May 2014.

[25]  Irene Kotsia, Iaonnis Pitas, "Facial expression recognition in image sequences using geometric deformation features and support vector machines", IEEE transaction paper on image processing, vol. 16, no.1, pp-172-187, January 2007.

[26] R.Rojas,"The back propagation algorithm",Springer-Verlag, Neural networks, pp 149-182 1996.

[27]  Hosseien Lari-Najaffi , Mohammad Naseerudin and  Tariq Samad,"Effects of initial weights on back propagation and its variations", Systems, Man and Cybernetics ,Conference Proceedings, IEEE International Conference, 14-17 Nov ,1989,Cambridge, pp-218-219.

[28]  M.H Ahmad Fadzil. , H.Abu Bakar., "Human face recognition using neural networks", Image processing, 1994, Proceedings ICIP-94, IEEE International Conference, 13-16 November, 1994, Austin , pp-936-939.

[29]  N.K Sinha , M.M Gupta and D.H Rao, "Dynamic Neural Networks -an overview", Industrial Technology 2000,Proceedings of IEEE International Conference,19-22 Jan,2000, pp-491-496.[30] Prachi Agarwal, Naveen Prakash, "An Efficient Back Propagation NeuralNetwork Based Face Recognition System Using Haar Wavelet Transform and PCA"   International Journal of Computer Science and Mobile Computing, vol.2, no.5,pg.386 – 395,May 2013.

[31] Dibber, 4 Jan 2005, "Backpropagation", https://en.wikipedia.org/wiki/Backpropagation, 20 September 2015

[32] "Artificial Neural Networks", https://en.wikipedia.org/wiki/Artificial_neural_network, accessed online 2 October 2001,

[33] Michael Nielsen Jan 2016, "Neural networks and deep learning", http://neuralnetworksanddeeplearning.com/chap2.html.[34] Tyrell turing, 7 April 2005, "Feed forward neural network", https://en.wikipedia.org/wiki/Feedforward_neural_network.